\documentclass[runningheads,a4paper]{llncs}

\usepackage{amssymb}
\usepackage{graphicx}
\usepackage{color}
\usepackage{floatrow}
\usepackage[T1]{fontenc}
\usepackage[font=small,labelfont=bf,tableposition=top]{caption}

\DeclareCaptionLabelFormat{andtable}{#1~#2  \&  \tablename~\thetable}

\usepackage{url}
\urldef{\mailsa}\path|{alfred.hofmann, ursula.barth, ingrid.haas, frank.holzwarth,|
\urldef{\mailsb}\path|anna.kramer, leonie.kunz, christine.reiss, nicole.sator,|
\urldef{\mailsc}\path|erika.siebert-cole, peter.strasser, lncs}@springer.com|

\begin{document}

\mainmatter  % start of an individual contribution
\pagenumbering{gobble}
% first the title is needed
\title{DAGER: Deep Age, Gender and Emotion Recognition Using Convolutional Neural Networks}

\author{Afshin Dehghan \hspace{0.25 in} Enrique G. Ortiz \hspace{0.25 in} Guang Shu \hspace{0.25 in} Syed Zain Masood \\
		{\tt\small \{afshindehghan, egortiz, guangshu, zainmasood\}@sighthound.com}}

% the affiliations are given next; don't give your e-mail address
% unless you accept that it will be published
\institute{Computer Vision Lab, Sighthound Inc., Winter Park, FL}

\maketitle

%%%%%%%%% ABSTRACT
\begin{abstract}
This paper describes the details of Sighthound's fully automated age, gender and emotion recognition system. The backbone of our system consists of several deep convolutional neural networks that are not only computationally inexpensive, but also provide state-of-the-art results on several competitive benchmarks. To power our novel deep networks, we collected large labeled datasets through a semi-supervised pipeline to reduce the annotation effort/time. We tested our system on several public benchmarks and report outstanding results. Our age, gender and emotion recognition models are available to developers through the Sighthound Cloud API at \textcolor{blue}{https://www.sighthound.com/products/cloud}
\end{abstract}

\section{Introduction}

Facial attribute recognition, including age, gender and emotion, \cite{emotion1,emotion2,emotion3,RotheICCV15,Gallagher2009,OrangeLab,WangWACV2015} has been a topic of interest among computer vision researchers for over a decade. One of the key reasons is the numerous applications of this challenging problem which range from security control, to person identification, to human-computer interaction.  Due to the release of large labeled datasets, as well as the advances made in the design of convolutional neural networks, error rates have dropped significantly. In many cases, these systems are able to outperform humans \cite{Gallagher2009}.  However, this still remains a difficult problem and existing commercial systems fall short when dealing with real world scenarios. In this work, we present an end-to-end system capable of estimating facial attributes including age, gender and emotion with low error rates. In order to support our claims, we tested our system on several benchmarks and achieved results better than the previous state-of-the-art. The contributions of this work are summarized below.

\begin{itemize}
\renewcommand{\labelitemi}{\scriptsize$\blacksquare$} 

\item We present an end-to-end pipeline, along with novel deep networks, that not only are computationally inexpensive, but also outperform competitive methods on several benchmarks.

\item We present large datasets for age, emotion and gender recognition that are used to train state-of-the-art deep neural networks.

\item We conducted a number of experiments on existing benchmarks and obtained leading results on all of them.

\end{itemize}
\section{System Overview}

\begin{figure}
\begin{center}
   \includegraphics[width=\linewidth]{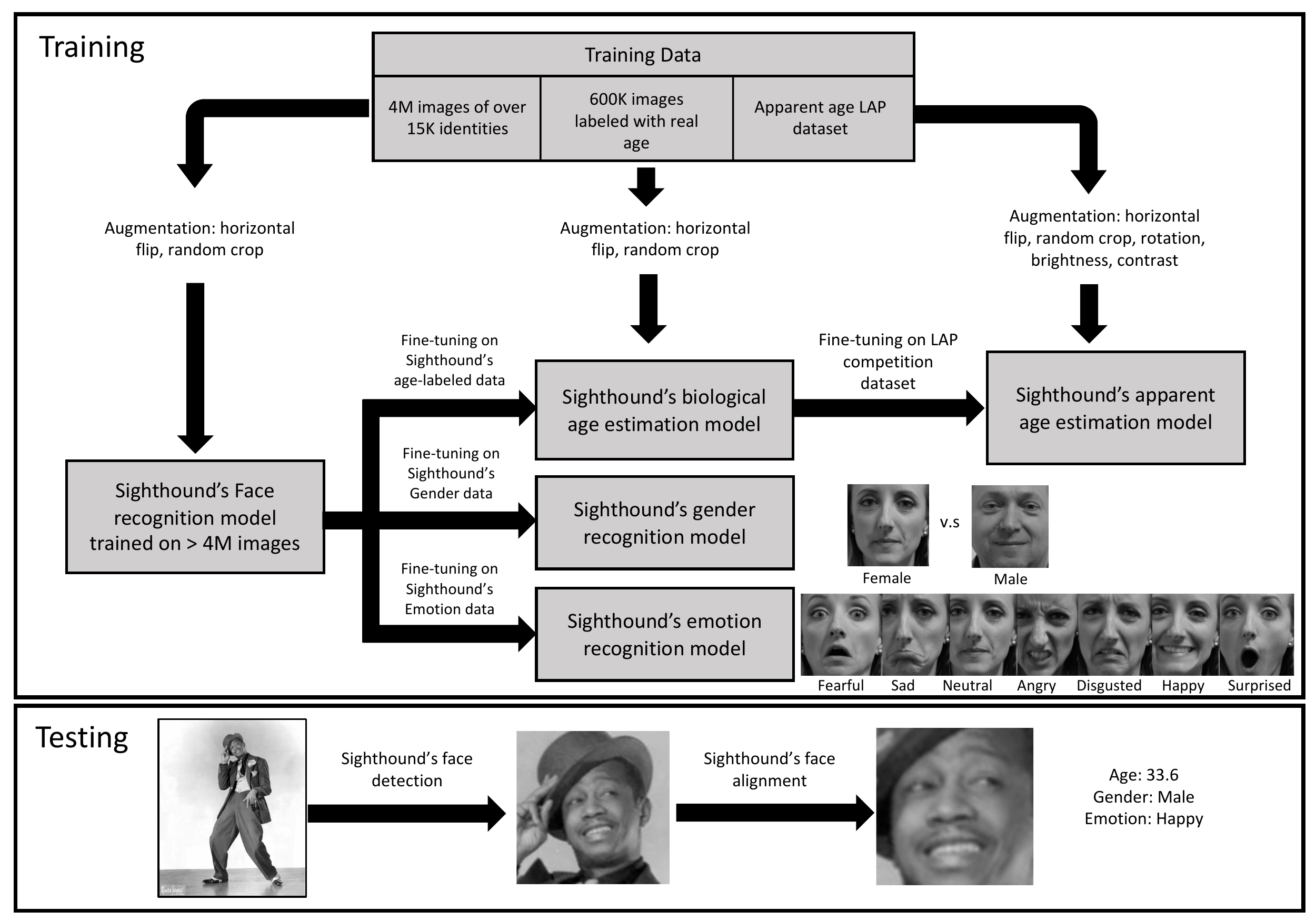}

   \caption{This figure shows the pipeline of our system. Images are collected from different sources and labeled for different tasks. Over 4 million images of more than 40,000 people are collected for the task of facial recognition. All images are labeled with their corresponding gender label and part of the data is annotated with emotion. These images are later pruned using a semi-automated process with a team of human annotators in the loop. The images are pre-processed next to extract the faces and align them. The aligned images are then fed to our proprietary deep network for training.}
   \label{figPipeline}

\end{center}
\end{figure}

The pipeline of our system is shown in Figure \ref{figPipeline}. Our first deep model is trained on a large dataset of four million images for the task of face recognition. This model serves as the backbone to our facial attribute recognizers and is used to fine-tune networks for four tasks: real age estimation, apparent age estimation, gender recognition and emotion recognition. What follows explains each of the steps in more detail.  

\subsection{Training}

Below we describe different steps involved in training our models in more detail. 

\begin{itemize}
\renewcommand{\labelitemi}{\scriptsize$\blacksquare$} 

\item{\textbf{Data collection}}: Data collection plays an important role in training any deep neural network (DNN). In this paper, we aim to label data for three separate tasks: age, gender and emotion recognition. Collecting labeled data for some tasks, such as real age estimation, is much more challenging compared to popular classification \cite{SighthoundMMCR,ImageNet} or detection \cite{PASCAL} problems. This disparity is due to the fact that human error in estimating real age is large (sometimes greater than the computer vision estimations) and one cannot rely on human annotators to label faces with their corresponding real age. However, at Sighthound we have collected a large dataset of faces with their corresponding age, gender and emotion labels. To our knowledge, our datasets are the largest or among the largest in either the academic or commercial world. Below we provide some statistics on the data used for training our models. 

\textbf{Face recognition:} The base model for our facial recognition is trained on over four million images of more than $40,000$ individuals. The large variation in images of each identity make our deep model robust to common challenges in face recognition. Our face recognition model is available to developers through the Sighthound Cloud API \footnote{https://www.sighthound.com/products/cloud}.

\textbf{Age estimation:} Recently there have been some efforts in collecting data with corresponding age labels \cite{Gallagher2009,Eidinger2013,RotheICCV15}. Among those, the dataset proposed by Rothe et al. in \cite{RotheICCV15} is the largest dataset that contains $523,051$ images and is available for research purposes. However, the dataset is not carefully annotated and contains many mistakes. Additionally the distribution of the data across different ages is highly unbalanced. This led to the authors using only half of the data for training in the original paper \cite{RotheICCV15}. To better address this problem we collected a large dataset of $\sim 600,000$ images with corresponding age labels. In contrast to previous works, our dataset has a more balanced distribution across different ages. For example we have over $120,000$ people in our dataset with labeled ages over $70$ or younger than $20$ years of age. We used a team of human annotators to further clean our dataset through a semi-supervised procedure.

\textbf{Gender and emotion recognition:} Our four million faces labeled for the task of face recognition are also labeled with their corresponding gender. To better improve our model, we added tens of thousands of images of different ethnicities as well as age groups. Additionally, we also annotated part of our data with emotion labels for the task of emotion recognition. \\

\item{\textbf{Data pre-processing}}:
We pre-process each image before feeding them to our DNNs. These pre-processing steps include face detection, facial landmark detection and alignment. We used Sighthound's face detection which is available through our cloud API. If more than one face is detected in an image, we choose the most centered one. (This is especially the case in the ChaLearn v2 dataset which contains multiple faces). In the Chalearn dataset we were able to detect all faces using a combination of techniques, but all using the Sighthound Cloud APIs.\footnote{Not all of the face detection bounding boxes generated by the cloud API were perfectly accurate, mostly due to occlusion or low resolution of some images. However, when comparing with other methods on pubic datasets, we directly used the output of our face detection for age estimation without further adjustment.} Given the face bounding boxes, we detect  $68$ facial landmarks and use those for alignment. Finally the aligned faces are all cropped and resized to a fixed size. In contrast to some previous works, which do not use any face alignment \cite{RotheICCV15}, we found this to be important in our final accuracy numbers. \\

\item{\textbf{Deep training}}: As shown in Figure \ref{figPipeline}, we start by training a deep neural network for the task of face recognition using four million images of over $40,000$ identities. Our face recognition model is not only computationally inexpensive (with feature extraction time of 70ms using just the CPU), but also achieves outstanding results on the LFW dataset [2]. This model serves as the backbone of our facial attribute recognition engine. We designed a highly optimized deep network architecture for accuracy and speed for each task. In some recent works \cite{allInOne}, researchers try to design a network which performs all tasks at the same time, and they have shown marginal improvements. However, having separate networks for each task allowed us to design faster and more portable models for each task. Additionally running all models combined takes less time compared to the all-in-one model of \cite{allInOne} and we achieve better results. We should add that even though the network architecture is not the same for each task, all networks are trained first for the task of facial recognition using the four million image set. 

\end{itemize}
\begin{figure}[t]
\begin{center}
\includegraphics[width=1\linewidth]{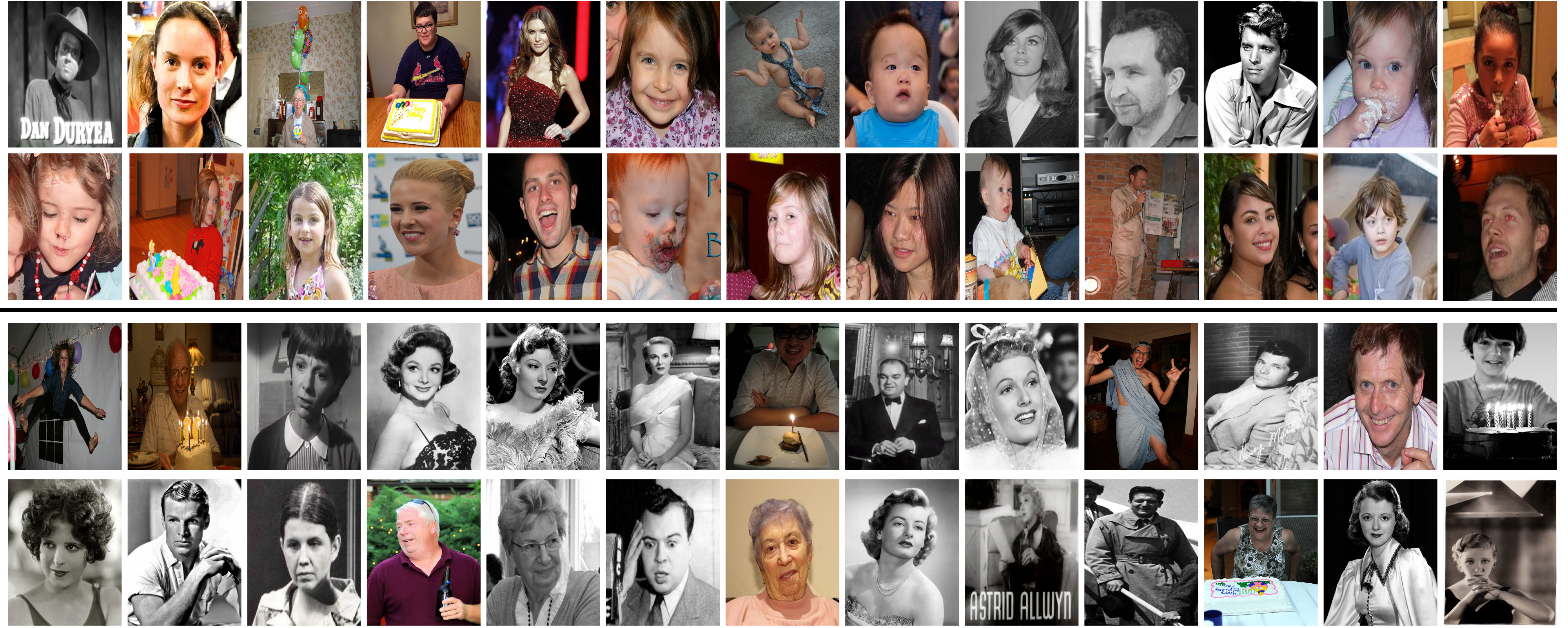}
\caption{Qualitative results of our method on ChaLearn LAP challenge. The top half of the figure shows some sample images where our absolute error is less than half a year and the bottom half shows images in which our error was more than $5$ years. As shown, the performance of our method drops mostly for gray-scale/old-style images where our network tends to over estimate the age for those images.}
\label{fig:emotion}
\end{center}
\end{figure}

\section{Experiments}
In this section, we report experimental results on several publicly available datasets as well as our internal datasets. 

\subsection{Real Age Estimation}

For real age estimation, we report results on two publicly available datasets, the Group dataset \cite{Gallagher2009} and the Adience dataset \cite{Eidinger2013}. The images in these datasets are labeled with their corresponding age groups. In order to further evaluate our system on estimating the actual age and not only the age group, we collected an internal dataset of $3,800$ images, on which we report the results of the proposed method in addition to other competitive algorithms.\\ 

\begin{table}[ht!]
  \caption{This table shows the mean absolute error for our methods along with competitive approaches on our Sighthound dataset. As can be seen, our method outperforms the second best method by $1.58$ years.}
  
  \centering
  \begin{tabular}{| @{\hspace{1em}}c@{\hspace{1em}} | @{\hspace{1em}}c@{\hspace{1em}} |} \hline
      Methods                        & MAE                         \\ \hline\hline
  \textbf{ \color{red}{Sighthound}}  & \color{red}\textbf{$5.76$}  \\ \hline    
   Rothe et al. \cite{RotheICCV15}   & $7.34$                      \\ \hline
   Microsoft. \cite{Microsoft}       & $7.62$                      \\ \hline
   Kairos \cite{Kairos}              & $10.57$                     \\ \hline
   Face++ \cite{FacePluSplus}        & $11.04$                     \\ \hline  
  \end{tabular}
  \label{tab:sighthound}
\end{table}

\vspace{+0.2in}
In Table \ref{tab:sighthound}, we present quantitative results on our Sighthound dataset, containing $3,800$ test images. Each image is labeled with its corresponding age ranging from $10$ to $90$ years old. Unlike the Adience and Group datasets, our dataset includes the exact age labels for each image and not the age groups. We compare our results with \cite{RotheICCV15}, whose model is trained on the IMDB-Wiki dataset. Additionally, we compare our system with available commercial APIs: Microsoft \cite{Microsoft}, Face++ \cite{FacePluSplus} and Kairos \cite{Kairos}. For quantitative comparison, we used the Mean Absolute Error (MAE) which is commonly used in the literature \cite{RotheICCV15}. As can be seen, our method outperforms competitive approaches by a significant margin.  \footnote{We compute the error rate for Microsoft and Face++ using the versions of their cloud API available in October 2016.} 

Next, we provide results on the Group dataset, which contains $28,231$ faces collected from Flickr and labeled with one of seven age categories roughly correspond to different life stages. Most faces are low-resolution making it more challenging for accurate age estimation. The median of faces are reported to have only $18.5$ pixels between the eye centers. We followed the setup in \cite{Gallagher2009} where $3,500$ images are used for training and $1,050$ images are used for testing. Both training and testing images are equally distributed across seven age groups. The age classification results in terms of accuracy along with a confusion table are reported in Table \ref{tab:group} and Figure \ref{fig:confMat_group}. We can observe that Sighthound's age estimation outperforms the latest research results.\\

  \begin{minipage}{\textwidth}
  \begin{minipage}[b]{0.49\textwidth}
    \centering
    \includegraphics[width=5cm,height=5cm,keepaspectratio]{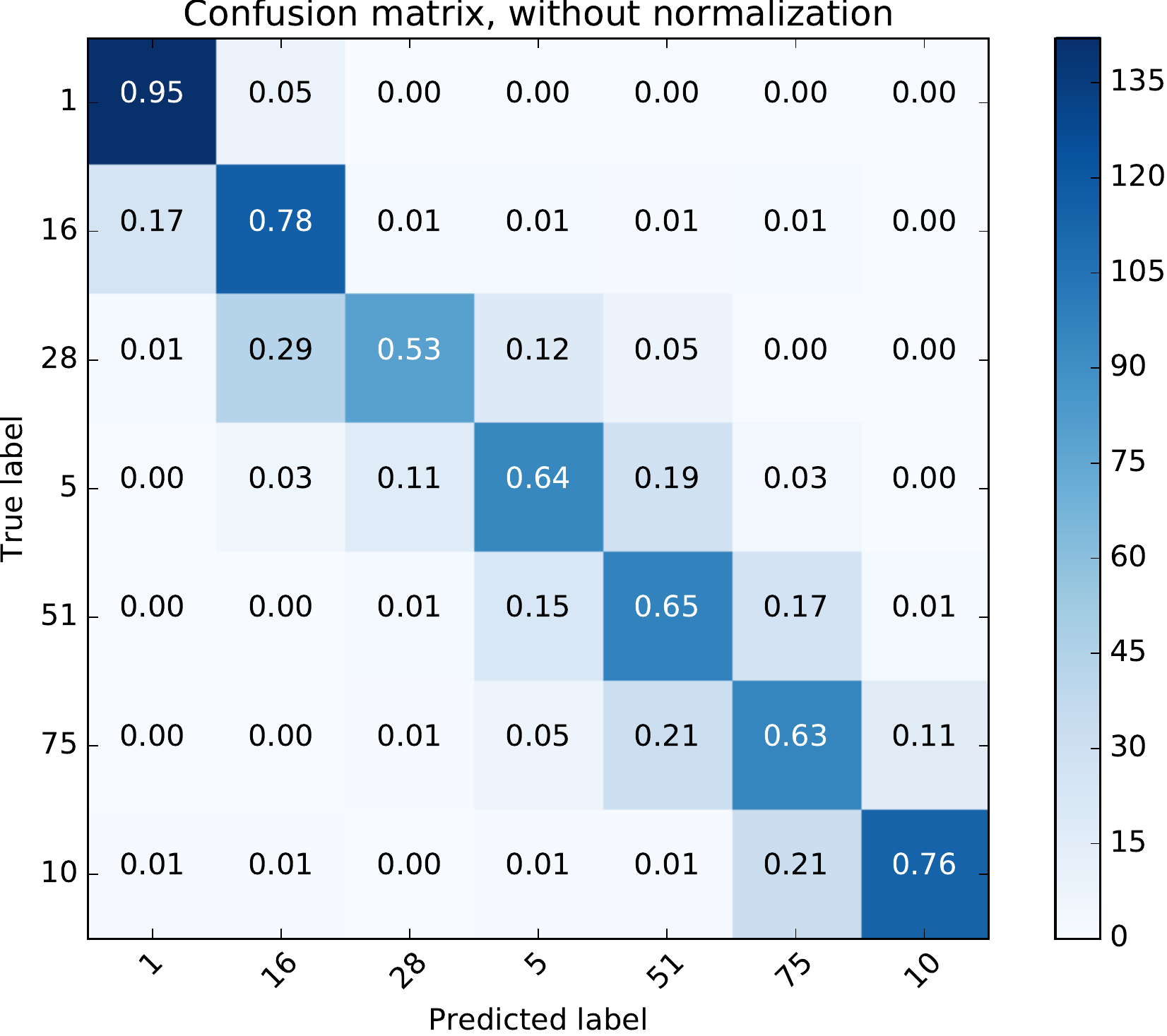}
    \captionof{figure}{Confusion table for the \newline Group dataset.}\label{fig:confMat_group}
  \end{minipage}
  \hfill
  \begin{minipage}[b]{0.49\textwidth}
    \centering
\begin{tabular}{|@{\hspace{0.2em}}c@{\hspace{0.2em}}|@{\hspace{0.2em}}c@{\hspace{0.2em}}|@{\hspace{0.2em}}c@{\hspace{0.2em}}|} \hline
Methods                        & top1                    & 1-off \\ \hline\hline
\textbf{ \color{red}{Sighthound}}      & \textbf{\color{red}{$70.5$\%}} & \textbf{ \color{red}{$96.2$\%}  }  \\ \hline
Hou et al. \cite{HouArxiv2016}        & $65.0$\%    & $96.1$\%        \\ \hline
Rothe et al. \cite{RotheICCV15}       & $62.3$\%    & $94.3$\%        \\ \hline
Dong et al. \cite{YuanNC2016}         & $56.0$\%    & $92.0$\%        \\ \hline
Gallegher et al. \cite{Gallagher2009} & $42.9$\%    & $78.1$\%        \\ \hline
\end{tabular}
  \vspace{0.2in}
   \captionof{table}{Age classification accuracy of the Group dataset. We report both the exact accuracy as well as the 1-off accuracy.}\label{tab:group}
   \end{minipage}
  \end{minipage}

\vspace{0.2in}
Finally we report results on the Adience benchmark. The entire Adience benchmark includes roughly $26,000$ images of $2,284$ subjects. However, some images are not annotated with corresponding age groups. Therefore the total number of images used for final testing is smaller than 26K. We used the standard 5-fold cross validation experiment defined for this set. When testing on each fold, the rest of the folds are used to fine-tune our model for the eight age groups defined in the dataset. The results of our method along with competitive approaches are shown in Table \ref{tab:adience}. Once again, our method improves on the best reported results on this dataset.

\begin{table}[ht!]
\caption{Real age estimation accuracy of the Adience benchmark. Sighthound outperforms other methods. (NR=Not Reported)}
\centering
\begin{tabular}{| @{\hspace{1em}}c@{\hspace{1em}} | @{\hspace{1em}}c @{\hspace{1em}}|}\hline
          Methods                      & Accuracy( Top-1 )   \\ \hline\hline
\textbf{ \color{red}{Sighthound}}      & \color{red}\textbf{$61.3\pm 3.7$\%}  \\ \hline    
Hou et. al. \cite{HouArxiv2016}        & $61.1\pm NR$\%            \\ \hline
Eidinger et. al. \cite{Eidinger2013}   & $45.1\pm2.6$\%      \\ \hline
Levi and Hassner \cite{Levi2015}       & $50.7\pm5.1$\%      \\ \hline

\end{tabular}
\label{tab:adience}
\end{table}

\vspace{-0.2in}
\subsection{Apparent Age Estimation}
The apparent age of a person could be very different from the real age of a person. Recently, thanks to the availability of the Chalearn LAP Apparent Age Estimation dataset and challenge \cite{Chalearn}, several researchers have focused on designing models that are focused on predicting the apparent age, rather than the actual age.  In the most recent version of the competition, the size of the dataset was extended to $7,591$ images where $4,113$ of them are used for training and $1,500$ and $1,978$ are used for validation and testing respectively.  Each image in the dataset is annotated using at least $10$ human votes and the mean ($\mu$) and standard devision ($\sigma$) of the votes are recorded and released with the dataset. Given the prediction for each image ($x$), the error for each image is computed using $\epsilon = 1-\exp - \frac{(\hat{x}-\mu)^2}{2\sigma^2}$. This means the apparent age on an image with small standard deviation gets penalized more compared to one with larger standard deviation. The winner of the competition \cite{OrangeLab} used a  multi CNN framework and achieved a  test error of $0.2411$.  However, this method \cite{OrangeLab} has several limitations.  Their network architecture is an order of magnitude slower in speed and an order of magnitude larger in size compared to ours. Additionally, the multiple CNNs (minimum of $88$ forward passes for each image) in their pipeline makes it impossible to use their system in a real-time or even close-to real-time application. We should also mention that all the runner up approaches suffer from the same limitations. Even though our goal is to keep the computational complexity low, we still achieve the outstanding error rate of $0.319$, which places our approach second. 

%The entire pipeline of our approach, including image pre-processing, takes roughly $110$ ms per image. The results are presented in Table \ref{tab:chalearn}.   

\begin{table}
  \caption{Results for ChaLearn \cite{Chalearn} apparent age estimation $2016$ challenge. Our fine-tuned system achieves a test error of $0.319$ and obtains the second best place. One should note that our model uses only a single CNN, which is not the case for most top performing teams. Additionally, our base model is almost an order of magnitude faster than the base CNN model of top performing teams (VGG).}% According to numbers reported in \cite{OrangeLab} and \cite{palm-seu} the processing time for each each is 6.5(s) and 10(s) respectively. On the other hand our method takes only a fraction of a second to run the pre-processing and age estimation for a single image.}
  \centering
 \begin{tabular}{|@{\hspace{1em}}c@{\hspace{1em}}|@{\hspace{1em}}c@{\hspace{1em}}|@{\hspace{1em}}c@{\hspace{1em}}|}
    \hline
   Methods         & Test Error          & Score-level fusion  \\ \hline\hline
  \textbf{ \color{red}{Sighthound}}& \color{red}\textbf{$0.319$}  & No  \\ \hline    
   OrangeLabs \cite{OrangeLab}  & $0.2411$                   & Yes  \\ \hline
   Palm-seu  \cite{palm-seu}                   & $0.3214$                   & Yes  \\ \hline
   CMP+ETH   \cite{Uricar-CVPRw-2016}       & $0.3361$             & Yes  \\ \hline
   WYU-CVL          & $0.3405$                   & No    \\ \hline
   ITU-SiMiT        & $0.3668$                   & Yes   \\ \hline
   Bogazici         & $0.3740$                   & Yes   \\ \hline
   MIPAL-SNU        & $0.4565$                   & Yes   \\ \hline

    \end{tabular}
\label{tab:chalearn}
\end{table}

\vspace{-0.4in}
\subsection{Emotion Recognition}

There are several public datasets for emotion recognition. FER-2013 \cite{FER2013} and EmotiW are among the popular ones. The FER dataset contains low-quality gray scale images of size $48 \times 48$ which is not very representative of real world scenarios. Access to the EmotiW dataset was not granted to us. Thus we collected our own dataset of $2,156$ images.  Each image is labeled with one of the $7$ labels of "happy", "sad", "neutral", "disgusted", "surprised", "fearful" and "angry". The data has a relatively equal distribution across the $7$ emotions. We compared our method with the Microsoft Face API \cite{Microsoft}. The results as well as the confusion tables are shown in Table \ref{tab:emotion} and Figure \ref{fig:emotion} respectively. As shown, Sighthound's emotion recognition system outperforms Microsoft by a $15\%$ margin. \footnote{Microsoft's API failed detecting a face in $193$ images. To be fair to Microsoft we removed these images while evaluating their method.}

\begin{table}[ht!]
  \caption{Emotion recognition accuracy on Sighthound dataset.}
  \centering
 \begin{tabular}{|@{\hspace{1em}}c@{\hspace{1em}}|@{\hspace{1em}}c@{\hspace{1em}}|}
    \hline
          Methods                  & Accuracy                   \\ \hline\hline
  \textbf{ \color{red}{Sighthound}}& \color{red}\textbf{$76.1\%$}  \\ \hline    
   Microsoft \cite{Microsoft}  & $61.3 \%$                         \\ \hline

    \end{tabular}
\label{tab:emotion}
\end{table}

\begin{figure}[ht!]
\begin{center}
\includegraphics[width=1\linewidth]{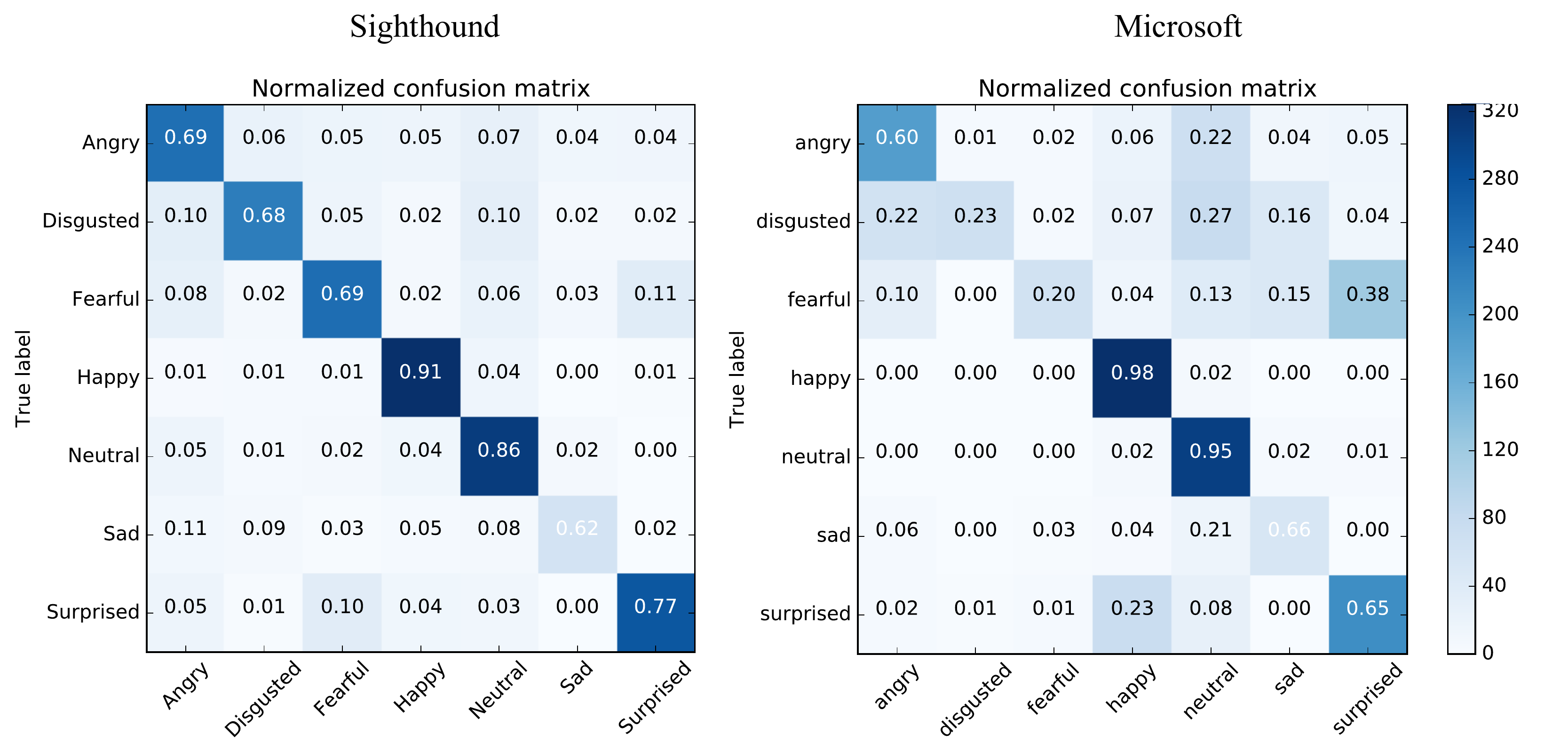}
\caption{Quantitative results in the form of a confusion matrix on Sighthound's emotion recognition dataset. On the left we show the results of our system and on the right we show the results for the Microsoft Face API. Our method performs well on almost all emotions, while Microsoft's performance drops significantly for emotions other than "happy" and "neutral".}
\label{fig:emotion}
\end{center}
\end{figure}

\subsection{Gender Recognition}

We compared our gender recognition model on the Adience benchmark with other leading methods. The Adience benchmark contains $17,492$ faces labeled with their corresponding gender. The faces are divided into $5$ folds. However, we used the same model across all folds without further fine-tuning. Along with published state-of-the-art results, we compare our method with a couple of commercial APIs such as \cite{FacePluSplus} and \cite{Kairos}. The results reported in Table \ref{tab:genderAdience} clearly show Sighthound's enhanced gender recognition capability compared to recent research publications and commercial products.

\begin{table}
  \caption{Results for gender recognition on the Adience benchmark \cite{Eidinger2013}. We compare our method against state-of-the-art research and commercial entities.}
  \centering
 \begin{tabular}{|@{\hspace{1em}}c@{\hspace{1em}}|@{\hspace{1em}}c@{\hspace{1em}}|}
    \hline
          Methods                     & Accuracy            		\\ \hline\hline
  \textbf{ \color{red}{Sighthound}}   & \color{red}\textbf{$91.00\%$}\\ \hline    
   Microsoft. \cite{Microsoft}    & $90.86\%$                  	\\ \hline
	Rothe et al. \cite{RotheICCV15}    & $88.75\%$                  	\\ \hline
   Levi and Hassner \cite{Levi2015}   & $86.80\%$                  	\\ \hline
   Kairos \cite{Kairos}               & $84.66\%$                   \\ \hline
   Face++ \cite{FacePluSplus}         & $83.04\%$                   \\ \hline
    \end{tabular}
\label{tab:genderAdience}
\end{table}

\section{Conclusions}

In this paper, we present an end to end system for age, gender and emotion recognition. We show that our novel deep architecture, along with our large, in-house collected data, can outperform competitive commercial and academic algorithms on several benchmarks.

\bibliographystyle{splncs}
\bibliography{egbib}

\begin{thebibliography}{10}

\bibitem{emotion1}
Busso, Carlos, e.a.:
\newblock Analysis of emotion recognition using facial expressions, speech and
  multimodal information.
\newblock In: Proceedings of the 6th international conference on Multimodal
  interfaces. (2004)

\bibitem{emotion2}
Levi, G., Hassner., T.:
\newblock Emotion recognition in the wild via convolutional neural networks and
  mapped binary patterns.
\newblock In: Proceedings of the 2015 ACM on International Conference on
  Multimodal Interaction. ACM. (2015)

\bibitem{emotion3}
Pang, L., Ngo., C.W.:
\newblock Mutlimodal learning with deep boltzmann machine for emotion
  prediction in user generated videos.
\newblock In: Proceedings of the 2015 ACM on International Conference on
  Multimodal Retrieval. ACM. (2015)

\bibitem{RotheICCV15}
Rothe, R., Radu~Timofte, L.V.G.:
\newblock Dex: Deep expectation of apparent age from a single image.
\newblock In: International Conference on Computer Vision (ICCV),. (2015)

\bibitem{Gallagher2009}
Gallagher, A.C., Chen., T.:
\newblock Understanding images of groups of people.
\newblock In: CVPR. (2009)

\bibitem{OrangeLab}
Antipov, Grigory, e.a.:
\newblock Apparent age estimation from face images combining general and
  children-specialized deep learning models.
\newblock In: Proceedings of the IEEE Conference on Computer Vision and Pattern
  Recognition Workshops. (2016)

\bibitem{WangWACV2015}
Wang, X., Guo, R., Kambhamettu, C.:
\newblock Deeply-learned feature for age estimation.
\newblock In: WACV. (2015)

\bibitem{SighthoundMMCR}
Dehghan, A., Masood, S.Z., Shu, G., Ortiz., E.G.:
\newblock View independent vehicle make, model and color recognition using
  convolutional neural network.
\newblock In: arXiv:1702.01721. (2017)

\bibitem{ImageNet}
Deng, Jia, e.a.:
\newblock Imagenet: A large-scale hierarchical image database.
\newblock In: Computer Vision and Pattern Recognition. (2009)

\bibitem{PASCAL}
Everingham, M., E.S.M.A.V.G.L.W.C.K.I.W.J., Zisserman, A.:
\newblock The pascal visual object classes challenge: A retrospective .
\newblock In: International Journal of Computer Vision. (2015)

\bibitem{Eidinger2013}
Eidinger, E., Enbar, R., Hassner., T.:
\newblock Age and gender estimation of unfiltered faces.
\newblock In: IEEE TRANSACTIONS ON INFORMATION FORENSICS AND SECURITY. (2013)

\bibitem{allInOne}
Ranjan, R., Sankaranarayanan, S., Castillo, C.D., Chellappa, R.:
\newblock An all-in-one convolutional neural network for face analysis.
\newblock In: arXiv:1611.00851. (2016)

\bibitem{Microsoft}
Microsoft-Face-API.:
\newblock
\newblock (https://www.microsoft.com/cognitive-services/en-us/face-api.)

\bibitem{Kairos}
Kairos.:
\newblock
\newblock (https://www.kairos.com/kairos-2.0/demos)

\bibitem{FacePluSplus}
Face++.:
\newblock
\newblock (http://old.faceplusplus.com/demo-detect/)

\bibitem{HouArxiv2016}
Hou, L., Yu, C.P., Samaras., D.:
\newblock Squared earth mover's distance-based loss for training deep neural
  networks.
\newblock In: arXiv. (2016)

\bibitem{YuanNC2016}
Yuan, D., Liu, Y., Lian., S.:
\newblock Automatic age estimation based on deep learning algorithm.
\newblock In: Neurocomputing. (2016)

\bibitem{Levi2015}
Levi, G., Hassner., T.:
\newblock Age and gender classification using convolutional neural networks.
\newblock In: CVPRW. (2015)

\bibitem{Chalearn}
Escalera, Sergio, e.a.:
\newblock Chalearn looking at people and faces of the world: Face analysis
  workshop and challenge 2016.
\newblock In: Proceedings of the IEEE Conference on Computer Vision and Pattern
  Recognition Workshops. (2016)

\bibitem{palm-seu}
Huo, Z., Yang, X., Xing, C., Zhou, Y., Hou, P., Lv, J., Geng, X.:
\newblock Deep age distribution learning for apparent age estimation..
\newblock In: IEEE Conference on Computer Vision and Pattern Recognition
  Workshops. (2016)

\bibitem{Uricar-CVPRw-2016}
U\v{r}i\v{c}\'{a}\v{r}, M., Timofte, R., Rothe, R., Matas, J., Gool, L.V.:
\newblock Structured output {SVM} prediction of apparent age, gender and smile
  from deep features.
\newblock In: Proceedings of IEEE conference on Computer Vision and Pattern
  Recognition Workshops, Las Vegas, USA (2016)

\bibitem{FER2013}
Goodfellow, I.J.:
\newblock Challenges in representation learning: A report on three machine
  learning contests.
\newblock In: International Conference on Neural Information Processing. (2013)

\end{thebibliography}

\end{document}